\documentclass[runningheads]{llncs}
\usepackage{graphicx}
\usepackage{tikz}
\usepackage{comment} 
\usepackage{amsmath,amssymb,amsfonts} % define this before the line numbering.
\usepackage{color}
\usepackage{booktabs}
\usepackage[symbol]{footmisc}

\newcommand{\etal}{{\em et al\,.}}       % et al.
\newcommand{\eg}{{e.g.}}           % e.g.
\newcommand{\ie}{{i.e.}}           % i.e.
\begin{document}
% \linenumbers
\pagestyle{headings}
\mainmatter
\def\ECCVSubNumber{429}  % Insert your submission number here

\title{Unsupervised Domain Attention Adaptation Network for Caricature Attribute Recognition} % Replace with your title

\begin{comment}
\titlerunning{ECCV-20 submission ID \ECCVSubNumber} 
\authorrunning{ECCV-20 submission ID \ECCVSubNumber} 
\author{Anonymous ECCV submission}
\institute{Paper ID \ECCVSubNumber}
\end{comment}
%******************

%\begin{comment}
\titlerunning{DAAN for Caricature Attribute Recognition}
\author{Wen Ji\inst{1}$^\dagger$
\and
Kelei He\inst{2,3}$^\dagger$* \and
Jing Huo\inst{1}*
\and
Zheng Gu\inst{1}
\and
Yang Gao\inst{1,3}
\email{\{hkl,huojing,gaoy\}@nju.edu.cn}\\
\email{\{jiwen,guzheng\}@smail.nju.edu.cn}
}

\authorrunning{W. Ji, K. He et al.}
\institute{State Key Laboratory for Novel Software Technology, Nanjing, China
\and
Medical School of Nanjing University, Nanjing, China
\and
National Institute of Healthcare Data Science at Nanjing University, Nanjing, China}
%\end{comment}
%******************
\maketitle
\footnotetext[2]{These authors contributed equally as co-first authors.}
\footnotetext[1]{Co-corresponding authors.}

\begin{abstract}
Caricature attributes provide distinctive facial features to help research in Psychology and Neuroscience. However, unlike the facial photo attribute datasets that have a quantity of annotated images, the annotations of caricature attributes are rare. To facility the research in attribute learning of caricatures, we propose a caricature attribute dataset, namely WebCariA. Moreover, to utilize models that trained by face attributes, we propose a novel unsupervised domain adaptation framework for cross-modality (\ie, photos to caricatures) attribute recognition, with an integrated inter- and intra-domain consistency learning scheme. Specifically, the inter-domain consistency learning scheme consisting an image-to-image translator to first fill the domain gap between photos and caricatures by generating intermediate image samples, and a label consistency learning module to align their semantic information. The intra-domain consistency learning scheme integrates the common feature consistency learning module with a novel attribute-aware attention-consistency learning module for a more efficient alignment. We did an extensive ablation study to show the effectiveness of the proposed method. And the proposed method also outperforms the state-of-the-art methods by a margin. The implementation of the proposed method is available at \url{https://github.com/KeleiHe/DAAN}.

\keywords{Unsupervised domain adaptation; Caricature; Attribute recognition; Attention}
\end{abstract}

\section{Introduction}
Caricatures are facial drawings of human faces with exaggerating facial features. 
Studying the latent information conveyed by caricatures has been long to the neurologists, psychologists, and also the computer scientists. The recognition of caricatures indicates the mechanism of human thoughts, and the knowledge learned by human during this task. Compared with photos, recognize the identities of caricatures may easier for humans  \cite{mauro1992caricature,perkins1975definition}. This indicates the most representative face features are not destroyed even the shape and appearance of faces have been largely changed. By contrast, they are usually harder for the machine learning methods to recognize and comprehend.

\begin{figure}[!t]
  \centering
  \includegraphics[width=\linewidth]{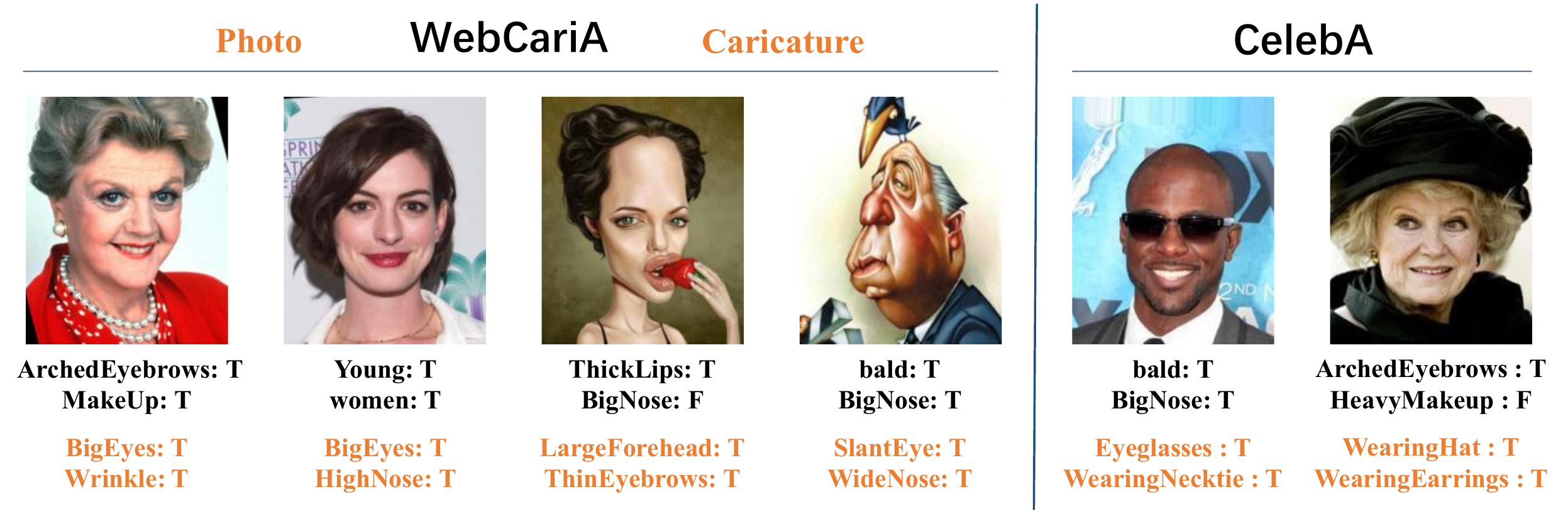}
    \caption{\label{fig:dataset} The comparison of typical cases in WebCariA and CelebA datasets. First two rows denote the same attributes annotated in the WebCariA dataset and the CelebA dataset. The last two rows indicate the distinctive attributes annotated in the two datasets. 'T' indicates 'True', 'F' indicates 'False'.}
\end{figure}

Recently, the research area of analyzing the caricatures in machine learning society has been raised, with several datasets are publicly released \cite{abaci2015matching,klare2012towards,huo2017webcaricature}. Most of the existing researches focus on face recognition \cite{mauro1992caricature,klare2012towards,huo2017webcaricature,valentine2016face} and image generation \cite{brennan2007caricature,cao2018carigans,kim2019u} using these datasets. However, as lacked by the annotations of the caricature attributes, a more valuable task of attribute recognition, has rarely been touched. To solve this problem, in this paper, we introduce the WebCariA dataset, by extending a large caricature dataset 'WebCaricture' \cite{huo2017webcaricature} with the annotation of fifty intrinsic face attributes. We hope it can boost this research area.
Specifically, to help understand the intrinsic facial characteristics that are felt by human, the face attributes on WebCariA are purely facial characteristics without the non-face attributes, compared with the existing face attribute in photo datasets (\eg, the attribute of 'Eyeglasses' and 'WearingHat' in CelebA dataset). (See the comparisons of typical examples in Fig. \ref{fig:dataset}) The details of the dataset will be further introduced in Section \ref{Sec:dataset}.

The face attribute recognition task has been solved well by deep convolutional networks \cite{liu2015deep,ding2018deep,NIPS20083499,smith2017federated,kumar2012learning}. It can be concluded as large-scale annotated face attribute datasets are already established, as the facial photos are easy to acquire. By contrast, the number of attribute annotated caricatures is small. Therefore, a natural idea is adapting a method that is trained on the annotated photos to the unannotated caricatures. This raises the problem of unsupervised domain adaptation from photos to caricatures for attribute recognition.

To solve the problem of cross-modality (\ie, face to caricature) attribute recognition on WebCariA dataset, we propose the domain attention adaptation network (DAAN), which has robust cross-domain adaptation ability for face attribute recognition. Specifically, to address the problem of large domain gap between photos and caricatures, DAAN has two main learning schemes to constrain the network, i.e., the inter-domain consistency learning and the intra-domain consistency learning. The inter-domain consistency learning scheme consisting of a pixel-level cross-domain image-to-image translator and the corresponding inter-domain consistency losses. The inter-domain consistency losses force the network to have consistent face attribute predictions on a certain image and its translated one. On the other hand, to align a certain image and a transferred image generated by the image of a different identity and domain property, we further propose the intra-domain consistency learning. It leverages both the feature and the attribute-aware attention map to build up the consistency between the two domains. Herein, we propose the attribute-aware attention-consistency learning by two observations: (1) the conventional feature-based consistency learning that align the distribution of the two domains in a global perspective is not efficient. (2) As a fine-grained classification task, face attributes are often revealed in a small region of the face, using attribute-aware attention will eliminate the noise conveyed by other parts. We did extensive ablation studies to show our proposed components, which can improve the discriminate ability of the network. The experiments also show our proposed method can outperform the state-of-the-art methods by a margin.

In this paper, our contributions are three-fold:
\begin{itemize}
    \item We introduce the WebCariA dataset with the annotation of fifty intrinsic face attributes on caricatures. 
    \item We propose a novel unsupervised attention adaptation framework for the recognition of attributes on unlabeled caricatures, which outperforms the state-of-the-art methods.
    \item We propose the attention-consistency learning which transfers the most task-discriminate features to achieve a more efficient adaptation.
\end{itemize}

\section{Related Work}
This work proposes an unsupervised domain adaptation method for attribute recognition in unpaired facial photos and caricatures. The related work can be concluded into three aspects: (1) The methods for face attribute recognition; (2) The methods for unsupervised domain adaptation in classification; (3) The generative adversarial learning-based image-to-image translation.

\subsection{Face Attribute Recognition Methods}

Face attribute recognition is a fine-grained classification task aiming to estimate facial characteristics. Previous works \cite{luo2012hierarchical,luo2013deep,zhu2014multi,liu2015deep,ding2018deep} based on convolutional neural networks have achieved satisfactory results. In the literature, early works mostly focus on single attribute learning \cite{geng2013facial,wang2015deeply}. After several large face attribute datasets being released, \eg, CelebA and LFWA \cite{liu2015deep}, more works attempt to study on multi-attribute estimation, formulating the inter-attribute relationships in one method. For example, Liu \etal \cite{liu2015deep} and Ding \etal\cite{ding2018deep} proposed to localize face regions for multiple attribute recognition. 
Typically, multi-task learning (MTL) with multiple classifiers is a common technique to boost the generalization ability of the method by exploiting the inter-attribute relationships. Han \etal proposed a deep MTL approach for multi-attribute estimation in heterogeneous faces. Lu \etal \cite{lu2017fully} learn a deep MTL framework that dynamically groups similar tasks together. Zhang \etal \cite{zhangposition} divide the attributes into different groups according to the location of the attribute. The multi-task learning have been proved to be very effective for facilitating the prediction of face attributes \cite{he2017adaptively,lu2017fully,wang2017multi,abdulnabi2015multi,ehrlich2016facial,rudd2016moon}.

\subsection{Unsupervised Domain Adaptation Methods in Classification}

The scenario of unsupervised domain adaptation (UDA) arises when we aiming at constructing a model that learns from an annotated data distribution (\ie, source domain), and need it to generalize well on a different (but related) unannotated data distribution (\ie, target domain).
Plenty of UDA methods for classification have been proposed in recent years. \cite{csurka2017domain} We can roughly divide them into two categories: 
(1) Non-adversarial learning methods: A metric is often used as the objective to directly measure and minimize the discrepancy of high-level features between the source domain and the target domain. For instance, the work in \cite{dan} uses the maximum mean discrepancy (MMD) as the metric. Vazquez \etal \cite{vazquez2012unsupervised} proposed to use a transductive SVM algorithm to solve the UDA problem. The non-adversarial methods have the ability to strongly align two different domains that is efficient for simple data distributions.
(2) Adversarial learning methods: The discrepancy between the source and the target domain is usually large. Therefore, the adversarial learning-based methods try to minimize such discrepancy with the learning of the domain distributions. For example, Ganin \etal \cite{dann} proposes a domain adversarial neural network (DANN) for image classification which contains a loss for label prediction and a loss for domain classification. The maximum classifier discrepancy (MCD) \cite{mcd} method builds up two classifiers to classify the source samples. Specifically, the two classifiers are forced to have different task-specific decision boundaries for the target sample. 
To make the method aware of a specific task, the Drop to Adapt (DTA) \cite{dta} method is proposed to learn robust and discriminate features by leveraging the adversarial dropout strategy, which supports the cluster assumption. However, these methods still cannot get satisfying performance as they only directly align the two domain distributions. Therefore, several works \cite{Russo_2018_CVPR,hoffman2018cycada} have adopted image-to-image translation into the UDA framework, with the assumption of the generated intermediate data helps domain alignment. Besides, previous methods align the features learned in two domains, that are redundant and not efficient.

\subsection{Image-to-Image Translation}

The goal of image-to-image translation is to learn the mapping between input images and output images. GAN has been widely utilized to solve the image-to-image translation problem, as it can learn the latent distribution of the images. For example, Isola \etal \cite{isola2017image} have proposed the conditional adversarial networks (conditional GAN) for the image-to-image translation problem. Zhu \etal \cite{zhu2017unpaired} used the cycle-consistent adversarial networks (CycleGAN) to tackle the problem without paired training examples. Kim \etal proposes a method namely U-GAT-IT \cite{kim2019u} which tried to combine the attention mechanism with the generative adversarial networks. It prompts the generator to focus on areas that specifically distinguish between the two domains, and the discriminator to focus on the difference between an original image and transferred image in the target domain.

\section{The WebCariA Dataset}
\label{Sec:dataset}
To date, several caricature datasets \cite{abaci2015matching,klare2012towards,huo2017webcaricature} are publicly available for caricature recognition. However, due to lack of caricature attribute annotations, there are little attempts to solve the task of caricature attribute recognition. 
In order to promote the research of caricature attribute recognition and generation, we construct a face attribute dataset, namely WebCariA, by labeling all images in the WebCaricature \cite{huo2017webcaricature} dataset. The dataset contains 6024 caricatures and 5974 photos of 252 people. Each image was labeled more than three times with fifty attributes of intrinsic face characters, and the final labels are determined by a voting strategy. The dataset is released at \footnote[1]{https://cs.nju.edu.cn/huojing/WebCariA.htm}.

Different from the photo dataset CelebA \cite{liu2015deep} and LFWA \cite{liu2015deep}, we did not mark the attributes that are changeable, such as 'WearingJewelry', 'WearingHats', etc. Instead, we labeled the attributes that can reflect the intrinsic characteristics of the faces such as 'BigNose', 'Bald', etc. (See typical cases compared to CelebA in Fig. \ref{fig:dataset}). Besides, we hope the WebCariA dataset can also promote the research area of caricature generation.

\section{Method}

In this section, we first provide the pipeline of our unsupervised domain adaptation framework. Secondly, we introduce the integrated-domain generalization learning paradigm which consists of the inter-domain consistency learning and the intra-domain consistency learning. Finally, we describe the multi-task learning setting for more precise face attribute recognition.

\begin{figure}[!t]
  \centering
  \includegraphics[width=\linewidth]{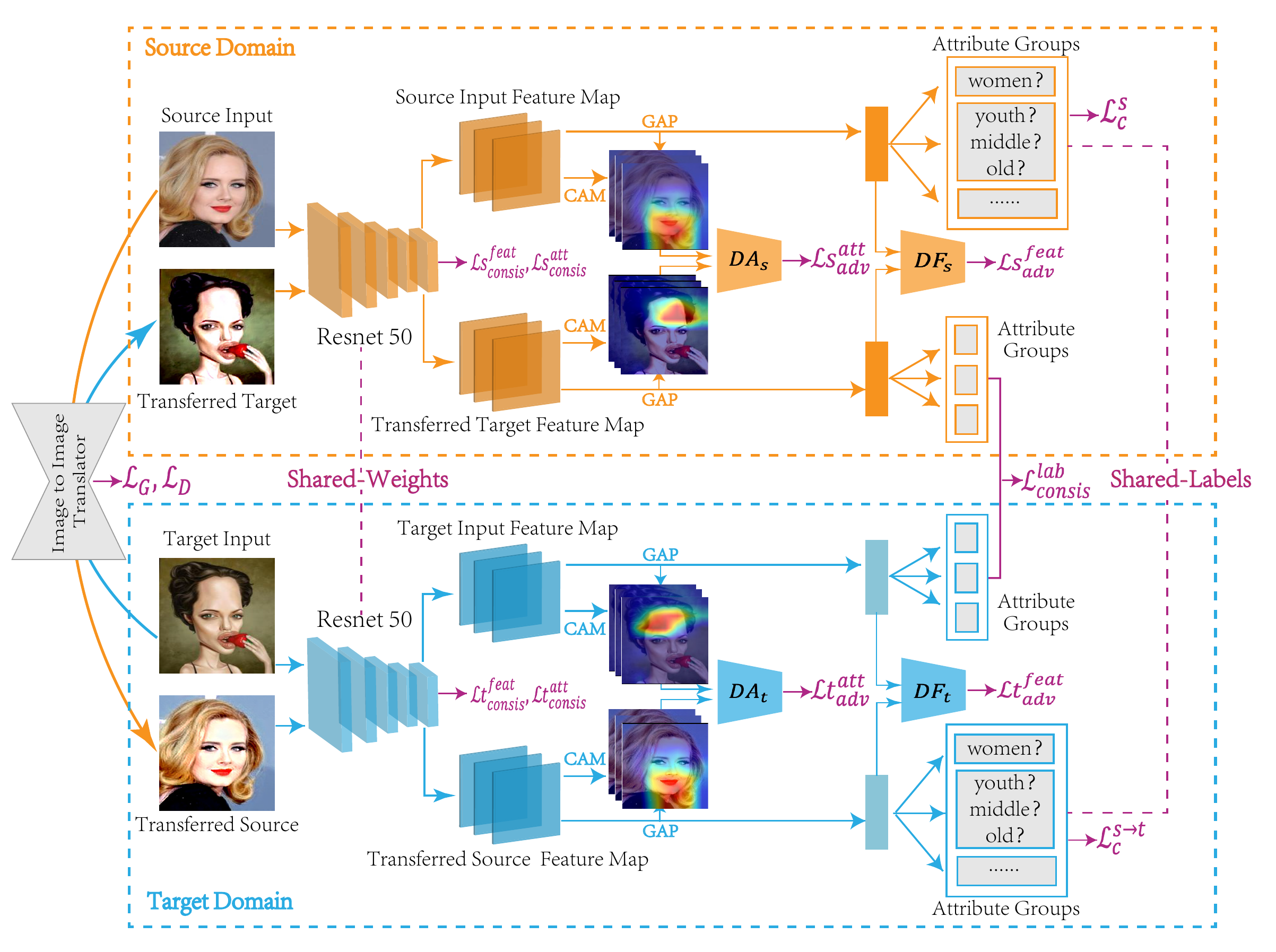}
    \caption{\label{fig:framework} The overall framework of our proposed unsupervised domain attention adaptation network (DAAN) for unsupervised face attribute recognition on caricatures.}
\end{figure}

\subsection{The Unsupervised Domain Adaptation Framework}

The overall framework is shown in Fig. \ref{fig:framework}. The goal of our method is to estimate face attributes on caricatures (\ie, the target domain) with the annotations of the attributes that are only given on photos (\ie, the source domain). Moreover, the images in source and target domain are unpaired, for which forms a typical unsupervised domain adaptation setting. Formally, let us denote the labeled source domain as $S = \left\{X_s, Y^i_s\right\}$, $i \in [1, N]$, where $\mathbf{x}_s \in X_s$ and $\mathbf{y}_s \in Y_s$ denote the source image and the corresponding label pair, $N$ denotes the number of face attributes. The unlabeled target domain can be therefore denoted as $T = \left\{X_t \right\}$, where $\mathbf{x}_t \in X_t$ denotes the target image. 

The domain gap and domain shift are the major problems for the cross-domain learning method. 
To solve the problems, our method adopted an integrated learning scene for learning both the inter- and intra-domain consistencies. The consistencies are learned by given constraints on the features. The inter-domain consistency learning firstly uses an unsupervised image-to-image translator to generate images from one domain to another by learning the influence of the style discrepancy of the two domains. The semantic information is preserved and the style information is transformed on the transferred images. Specifically, in this work, the image translation is bi-directional, where each image of the source domain will generate one transferred image in the target domain, and vise versa. The bi-directional image-to-image translation module in the framework can be regarded as to help build an intermediate data distribution between the source data and target data.

However, caricatures and photos have a large domain gap, \ie, with large style and shape variations. To constrain the semantic information conveyed by image and its generated one align in features, we introduce the label-consistency learning to minimize the prediction error generated on the two domains. 

Obviously, the generated data (from the other domain) and the real data in a certain domain have smaller domain gap, and thus are easier to be aligned together. We further propose an attention adaptation followed by a feature adaptation learning scheme to close the features of the intra-domain images, \ie, the original image and the transferred image (from another domain). 

For inference, we use a multi-task learning method to design the face attribute classifiers. By assuming that the attributes have mutually exclusions, we build attribute-groups to involve mutual attribute competitions. The final network is end-to-end, and all the modules we introduced can be trained simultaneously.

\subsection{Inter-Domain Consistency Learning}

\subsubsection{Unsupervised image-to-image translation.}

In the cross-domain attribute recognition task, the distribution between the source and target domains are large. 
Photos are shots of the real-world that often contain rich illumination, texture and noise. And caricatures are facial drawings are often concise, with clear outlines and exaggerated facial features. In order to eliminate the domain gap between the photo and caricatures, we first utilize an unsupervised image-to-image translator to alleviate the style discrepancy between these two domains. The translator takes a source image $\mathbf{x}_s$ and a target image $\mathbf{x}_t$ as inputs, and translate them from one domain to the other. The obtained corresponding translated images are denoted as $\mathbf{x}_{s\to t}$ and $\mathbf{x}_{t\to s}$.
Here we use the generator and discriminator proposed in \cite{kim2019u} as the image-to-image translator $\mathbf{G}_{s\to t}$ and $\mathbf{G}_{t\to s}$. Therefore, the loss are denoted as $\mathcal{L}_G$, $\mathcal{L}_D$, respectively.
In this way, the style difference between the source domain and the target domain is eliminated, making their domain distribution more consistent with each other. 
Then, the features of the four images are extracted by a feature extractor. The feature extractor $Resnet$ takes $\mathbf{x}_s$ and $\mathbf{x}_{t\to s}$ as inputs, then the feature vectors $\mathbf{F}_s$ and $\mathbf{F}_{t\to s}$ is obtained after the global average pooling (GAP). The features $\mathbf{F}_t$ and $\mathbf{F}_{s\to t}$ in the target domain are acquired accordingly. 

\subsubsection{Label-consistency learning.}

By assuming the semantic information is not destroyed during image translation, we give $\mathbf{x}_s$ and $\mathbf{x}_{s\to t}$ the same labels to calculate the classification loss. Here, we use the cross-entropy loss to calculate the attribute estimating errors with the input of $\mathbf{x}_s$ and $\mathbf{x}_{s\to t}$,

\begin{align}
  \mathcal{L}_c^s = -\mathbb{E}_{\mathbf{x}_s, \mathbf{y}_s\sim S}\left[\mathbf{y}_s^\top \log C\left(Resnet\left(\mathbf{x}_s \right) \right) \right] \\
  \mathcal{L}_c^{s\to t} = -\mathbb{E}_{\mathbf{x}_{s\to t}, \mathbf{y}_s\sim T'}\left[\mathbf{y}_s^\top \log C\left(Resnet\left(\mathbf{x}_{s\to t} \right) \right) \right]
\end{align}

where $C$ is the classifier, $S'$ and $T'$ are the distributions of the transferred image features. This also helps to make full use of annotations given in the source domain for semantic attribute consistency. 

On the other hand, the labels of $\mathbf{x}_t$ and $\mathbf{x}_{t\to s}$ should be predicted as consistent as possible, therefore, we design the label-consistency loss for $\mathbf{x}_t$ and $\mathbf{x}_{t\to s}$ to control semantic attribute consistency,

\begin{align}
    \mathcal{L}^{lab}_{consis} = -\mathbb{E}_{\mathbf{x}_t\sim T, \mathbf{x}_{t\to s}\sim S^\prime}\left[\left\|C\left(Resnet\left(\mathbf{x}_t \right) \right) - C\left(Resnet\left(\mathbf{x}_{t\to s} \right) \right) \right\|_2\right] \\
    \mathcal{L}_{inter} = \lambda_G\mathcal{L}_G + \lambda_D\mathcal{L}_D + \lambda_l\mathcal{L}^{lab}_{consis}
\end{align}

The proposed method achieves style consistency and semantic consistency between the source domain and the target domain by adopting the above-mentioned techniques. However, attribute recognition is a fine-grained classification task, these consistencies are built on a global perspective. They are not sensitive to the tiny differences between the two domains in the cross-domain face attribute recognition task. And the network still lack of the constraints between the transferred image and original image in one domain, \eg, $\mathbf{x_{s\to t}}$ and $\mathbf{x_t}$. We propose the intra-domain consistency learning to solve the problem.

\subsection{Intra-Domain Consistency Learning}

To make the source (/target) domain image and the transferred target (/source) image have stronger consistency constraints, we use the generative adversarial strategy to make the distribution of features align between the source domain and target domain. 
We introduce two discriminators $D_{Fs}$ and $D_{Ft}$ for the output features of the two domains, to construct the feature-level domain adaptation. 

Under the adversarial learning setting, the loss $\mathcal{L}^{feat}_{consis}$ of the features can be written as,

\begin{align}
  \mathcal{L}^{feat}_{consis} = \mathbb{E}\left[\sum \mathbf{Y}_s \log\left(D_{Fs}\left(\mathbf{F}_{t\to s}\right) \right)\right] + \mathbb{E}\left[\sum \mathbf{Y}_{s\to t} \log\left(D_{Ft}\left(\mathbf{F}_t \right)\right) \right]
  \label{Eq:Consis}
\end{align}

where $\mathbf{Y}$ is the originated domain label of the samples, and have the same shape to $\mathbf{F}$. If the samples come from source domain the elements of Y are 1 else 0. As the two pair of cross-domain features are jointly optimized, the final consistency loss in Eq. \ref{Eq:Consis} aggregate the errors raised by the two features. And in the discriminate process, the intra-domain feature consistency for source domain try to achieve feature-level alignment between $\mathbf{F}_s$ and $\mathbf{F}_{t\to s}$, which can be defined as follows,

\begin{align}
  \mathcal{L}s^{feat}_{adv} = \mathbb{E}\left[\log\left(D_{Fs}\left(\mathbf{F}_s \right) \right)\right] + \mathbb{E}\left[\log\left(1 - D_{Fs}\left(\mathbf{F}_{t\to s} \right) \right) \right]
\end{align}

For the features in the target domain, the alignment can be written as,

\begin{align}
  \mathcal{L}t^{feat}_{adv} = \mathbb{E}\left[\log\left(D_{Ft}\left(\mathbf{F}_{s\to t} \right) \right)\right] + \mathbb{E}\left[\log\left(1 - D_{Ft}\left(\mathbf{F}_t \right) \right) \right]
\end{align}

The features convey the whole information of an image, that is rough and not efficient. It is proved that discriminative features often make more contribution to the classification task \cite{vaswani2017attention}. And a slight difference between the source and target domain can seriously affect the classification results in face attributes classification,  because it is a fine-grained classification problem. 
Moreover, the distortion and exaggerations of caricature are often beyond realism which makes the gap of photos and caricatures very large.
Therefore, we propose the attribute-attention alignment between the original images and translated images (from the other domain). The attention maps $\mathbf{A}_s$, $\mathbf{A}_{t\to s}$, $\mathbf{A}_t$ and $\mathbf{A}_{s\to t}$ of all inputs are calculated by class activation map (CAM) \cite{zhou2016learning}. Take the source original image as example, $\mathbf{A}_s^i=\sum_{j=1}^{n}{w_j^i \mathbf{F}_{s,j}^\prime}$, where $\mathbf{F}_{s,j}^\prime$ is the $j$th channel of output feature maps $\mathbf{F}_s^\prime$ which is a metric before GAP and $w_j^i$ is the $j$th weight of the fully connected layer for predicting the $i$th category. Similar to the feature domain adaptation, we propose two attention-based discriminators, \ie, $D_{As}$ which is performed on $\mathbf{A}_s$ and $\mathbf{A}_{t\to s}$, and $D_{At}$ which is performed on $\mathbf{A}_t$ and $\mathbf{A}_{s\to t}$ to construct the attribute-attention domain adaptation.

The attribute attention-based adaptation can locate the decisive image region for classification. And with the help of the discriminators, the network will pay more attention to the inconsistent regions of the two domains to eliminate their tiny differences. Therefore, attention is suitable for domain alignment in face attributes classification. 

The attention-consistency loss can be defined as follow:
\begin{align}
  \mathcal{L}^{att}_{consis} = \mathbb{E}\left[\sum \mathbf{Y^\prime}_s \log\left(D_{As}\left(\mathbf{A}_{t\to s} \right) \right)\right] + \mathbb{E}\left[\sum \mathbf{Y^\prime}_{s\to t} \log\left(D_{At}\left(\mathbf{A}_t \right)\right) \right],
\end{align}

where $\mathbf{Y^\prime}$ is the label of the domain where the sample is coming from, its shape is same to $\mathbf{A}$. If the samples are drawn from the source domain, the elements of $\mathbf{Y^\prime}$ is 1 else 0. And the attention-based intra-domain consistencies $\mathbf{A}_s$ and $\mathbf{A}_{t\to s}$ for the source domain are defined as,

\begin{align}
  \mathcal{L}s^{att}_{adv} = \mathbb{E}\left[\log\left(D_{As}\left(\mathbf{A}_s \right) \right)\right] + \mathbb{E}\left[\log\left(1 - D_{As}\left(\mathbf{A}_{t\to s} \right) \right) \right]
\end{align}

Similarly, $D_{At}$ aligns the distributions between $\mathbf{A}_t$ and $\mathbf{A}_{s\to t}$,
\begin{align}
  \mathcal{L}t^{att}_{adv} = \mathbb{E}\left[\log\left(D_{At}\left(\mathbf{A}_{s\to t} \right) \right)\right] + \mathbb{E}\left[\log\left(1 - D_{At}\left(\mathbf{A}_t \right) \right) \right]
\end{align}

Overall, the final intra-domain consistency loss can be written as,
\begin{align}
    \mathcal{L}_{intra} = \lambda_f(\mathcal{L}s^{feat}_{consis}+\mathcal{L}t^{feat}_{consis})+\mathcal{L}s^{feat}_{adv}+\mathcal{L}t^{feat}_{adv}\\+\lambda_a(\mathcal{L}s^{att}_{consis}+\mathcal{L}t^{att}_{consis})+\mathcal{L}s^{att}_{adv}+\mathcal{L}t^{att}_{adv}
\end{align}

\subsection{Multi-Task Attribute Recognition}

For the task of caricature attribute recognition, the attributes are not independent. It is not suitable to treat them as a pure multi-class classification problem, as done by the conventional attribute recognition methods. Thus, the work in  \cite{zhangposition} partitioned all the attributes into different groups based on the global and local spatial regions. 

The idea is natural but not very reasonable to the learning theory. By contrast, we observe that in most face attribute datasets, there are mutually exclusive relationships among the face attributes. For example, the attributes of young, middle-aged and old are related to the group of age. Obviously, the mutually exclusive relationships among the attributes provides stronger and more stable constraints to build up the classifiers.
Herein, for the classifiers, we use one fully-connected layer to map the features to a certain number of classes for each attribute group. The cross-entropy loss is used to calculate the error. The results of each classifier is concatenated to obtain the final prediction result. 

Finally, the objective function involving all the previous mentioned losses can be written as,
\begin{align}
  \mathcal{L} = \mathcal{L}_c^s+\mathcal{L}_c^{s\to t}+\mathcal{L}_{inter}+\mathcal{L}_{intra}
\end{align}

\subsection{Implementation Details}

We implement the proposed method using the open-source framework \emph{PyTorch}. We use ResNet-50 \cite{he2016deep} without the layer after the global average pooling layer as the feature extractor, and share the extractor through all inputs in this work. The discriminators are randomly initialized with a structure similar to \cite{tsai2018learning}, consisting five convolutional layers with size of \{$3\times3$,1,1\} for \{kernel, stride, padding\}. We use the Stochastic Gradient Descent (SGD) algorithm to optimize the network, with a batch size of $40$. The learning rate for the feature extractor and the classifiers is $0.05$, with momentum of $0.9$ and weight decay of $5\times10^{-4}$. The learning rate is decayed by the polynomial strategy \cite{chen2017deeplab} with the power of $0.75$. For the discriminators, we use Adam optimizer with the learning rate of $1\times10^{-4}$. For the adversarial losses, we use the learning rates of $0.02$, $0.1$ and $0.1$ for $\lambda_l$, $\lambda_f$ and $\lambda_a$, respectively.

\section{Experimental Results}

Because annotated photos are often easy to acquire, in this paper, the method is built to estimate caricature attributes by the annotated photos, and using the photos in WebCariA dataset as the source domain. We use the metrics of accuracy (Acc.) and F1 score to measure the performance through the experiments. Please note that the attribute recognition task is very task-imbalanced for each attribute, the F1 score is more representative to show the real performance of the methods.

\subsection{Ablation Study}

We did extensive experiments for DAAN, including (1) the evaluation of the effectiveness of the MTL method, (2) the performance comparison of different network configurations of DAAN. The parameter analysis of DAAN is reported in \emph{Supplementary Material}.

Firstly, we construct the 'Source Only' method, which is only trained on the source domain and evaluate on the target domain; and the 'Target Only' method, which is trained on the target domain and also evaluates on the target domain, forms a fully-supervised fashion. These two methods are usually used as the baselines to indicate the lower and upper bound of the performance of domain adaptation. For the proposed DAAN, we construct different configurations of DAAN, with or without the consistency learning of label, feature and attribute-aware attention.

\begin{table}[!htbp]
\begin{center}
\setlength{\tabcolsep}{5mm}{
\centering
\caption{\label{Table:MTL} The effectiveness of the proposed multi-task learning strategy. 'MultiTask' denotes the network is learned with the attribute groups.}
\begin{tabular}{ccc}
\toprule[1pt]
\textbf{Method} & Avg. Acc  &  Avg. F1 \\
\toprule[1pt]
\textbf{Source Only} & 0.8054 & 0.6770 \\
\textbf{Source Only(MultiTask)} & 0.8050 & 0.6922 \\
\hline
\textbf{DTA \cite{dta}} & 0.8100 & 0.6941 \\
\textbf{DTA(MultiTask) \cite{dta}} & 0.8076 & 0.7000 \\
\hline
\textbf{Target Only} & 0.8474 & 0.7358 \\
\textbf{Target Only(MultiTask)} & 0.8526 & 0.7601 \\
\toprule[1pt]
\end{tabular}
}
\end{center}
\end{table}

\subsubsection{The effectiveness of multi-task learning.}

Table \ref{Table:MTL} shows the performance in average accuracy and average F1 score through all attributions on the WebCariA dataset. The table suggests that using the multi-task learning strategy consistently improves the performance of the network. For the two baseline networks 'Source Only' and 'Target Only' which are purely unsupervised and supervised methods, group attributes with multi-task learning improve the average F1 score by 1.52\% and 2.43\%, respectively. To evaluate the generalization of the proposed MTL, we implement the state-of-the-art unsupervised domain adaptation method in \cite{dta} with a single classifier, denote as DTA, and with the proposed MTL, denote as DTA(MultiTask). The experiments show that, for the well designed unsupervised domain adaptation method, our proposed MTL still improves the F1 score by 0.59\%.

\begin{table}[!htbp]
\setlength{\tabcolsep}{3mm}
\centering
\caption{\label{Table:Effect} Performance comparison with different configurations of DAAN. (Bests are in Bold)}
\begin{tabular}{clccccc}
\toprule[1pt]
\multicolumn{2}{c}{\textbf{Method}} & label & feat. & att. & Avg. Acc  &  Avg. F1 \\
\toprule[1pt]
\textbf{DAAN}&\textbf{-L} &\checkmark &&& 0.8122 & 0.7038 \\
&\textbf{-F} &&\checkmark&& 0.8169 & 0.7094 \\
&\textbf{-A} &  &&\checkmark& 0.8147 & 0.7107 \\
&\textbf{-LF} &\checkmark&\checkmark&& 0.8219 & 0.7181 \\
&\textbf{-LA} &\checkmark & &\checkmark& 0.8212 & 0.7192 \\
&\textbf{-LFA} &\checkmark &\checkmark&\checkmark & \textbf{0.8239} & \textbf{0.7215} \\
\toprule[1pt]
\end{tabular}
\end{table}

\subsubsection{Evaluation of different adaptation constraints.}

To evaluate the effectiveness of different adaptation constraints proposed by the method, we first construct DAAN with only one consistency constraint of label, feature and attribute-aware attention as 'DAAN-L','DAAN-F' and 'DAAN-A', respectively. Then, we compose two and three constraints into DAAN. The evaluations in average accuracy and F1 score are reported in \ref{Table:Effect}. We make three conclusions according to the table: (1) All the constraints help to regularize the features under the UDA setting. (2) Among one factor constrained DAANs, the attention-consistency constraint performs better than the other two constraints, indicates the attention is more efficient. (3) The combination of three constraints can further improve the model, and achieve the overall best performance. This reveals multiple constraints are useful to the UDA setting, which has more freedom in parameter space compared with fully-supervised learning.

\subsection{Comparison with the State-of-the-art Methods.}

\begin{figure}[!t]
  \centering
  \includegraphics[width=\linewidth]{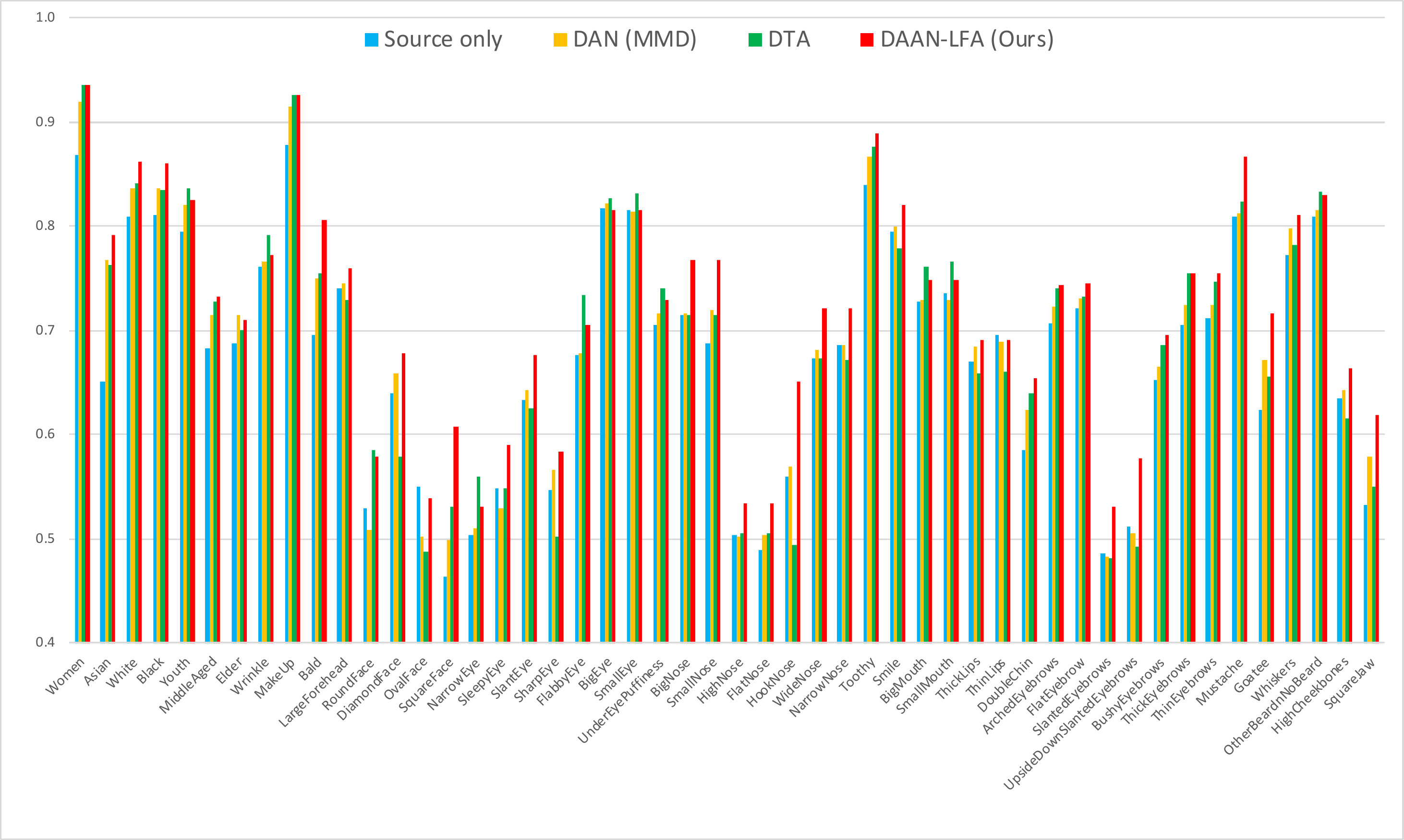}
    \caption{\label{fig:attRes} The performance comparison in F1 score with Source Only, DAN (MMD), DTA and the proposed DAAN-LFA.}
\end{figure}

We compare DAAN with several recent state-of-the-art methods for image classification. The compared methods include DAN \cite{dan}, DANN \cite{dann}, MCD \cite{mcd} and DTA \cite{dta}. The compared methods are reimplemented on the WebCariA dataset without our proposed MTL by mutually exclusive attribute groups. The performance in average accuracy and F1 score is reported in Table \ref{Table:sota}, and the attribute-wise performance in the F1 score is illustrated in Fig. \ref{fig:attRes}. The table suggests that the four state-of-the-art methods can both improve the performance of the network when adapting to another domain, and DTA performs best among the four methods in the second set of rows. So we further implement it with our MTL setting (denoted as 'DTA(MultiTask)'). As shown by the table, compared DAAN-LF with DTA, our proposed method gets a 1.19\% improvement in average accuracy and a 2.4\% improvement in average F1 score. Compared with DAAN-LFA with DTA and DTA(MultiTask), the improvements are 1.39\%, 1.63\% in average accuracy and 2.74\%, 2.15\% in average F1 score, respectively. This shows the effectiveness of DAAN compared with the current state-of-the-art methods. Furthermore, as shown in Fig. \ref{fig:attRes}, DAAN-LFA performs consistently best in most attributes, when compared with Source Only, DAN and DTA method. It is worth noting that DAAN performs especially well in attributes that are hard to estimate (\ie, with F1 score under 0.6), \eg, SquareFace, HookNose, etc. This demonstrates the robustness of the proposed DAAN. We also analyze the generalization ability for DAAN on small benchmark datasets, which is reported in \emph{Supplementary Material}.

\begin{table}[!h]
\begin{center}
\setlength{\tabcolsep}{5mm}{
\centering
\caption{\label{Table:sota} Comparison with the state-of-the-art methods for performance in average Acc. and F1 score. As MCD method proposed two classifiers in the network, we report both of them separately.}
\begin{tabular}{ccc}
\toprule[1pt]
\textbf{Method} & Avg. Acc  &  Avg. F1 \\
\toprule[1pt]
\textbf{Source Only} & 0.8054 & 0.6770 \\

\textbf{Source Only(MultiTask)} & 0.8050 & 0.6922 \\
\hline
\textbf{DAN \cite{dan}} & 0.8098 & 0.6921 \\
\textbf{DANN \cite{dann}} & 0.7783 & 0.6604 \\
\textbf{MCD \cite{mcd}} & 0.7967 (0.7993) & 0.6920 (0.6914) \\
\textbf{DTA \cite{dta}} & 0.8100 & 0.6941 \\
\hline
\textbf{DTA(MultiTask) \cite{dta}} & 0.8076 & 0.7000 \\
\hline
\textbf{DAAN-LF(Ours)} & 0.8219 & 0.7181 \\
\textbf{DAAN-LA(Ours)} & 0.8212 & 0.7192 \\
\textbf{DAAN-LFA(Ours)} & \textbf{0.8239} & \textbf{0.7215} \\
\hline
\textbf{Target Only} & 0.8474 & 0.7358 \\
\textbf{Target Only(MultiTask)} & 0.8526 & 0.7601 \\
\toprule[1pt]
\end{tabular}
}
\end{center}
\end{table}

\subsection{Visualization of the Attribute-wise Attention Maps}
\begin{figure}[!t]
  \centering
  \includegraphics[width=\linewidth]{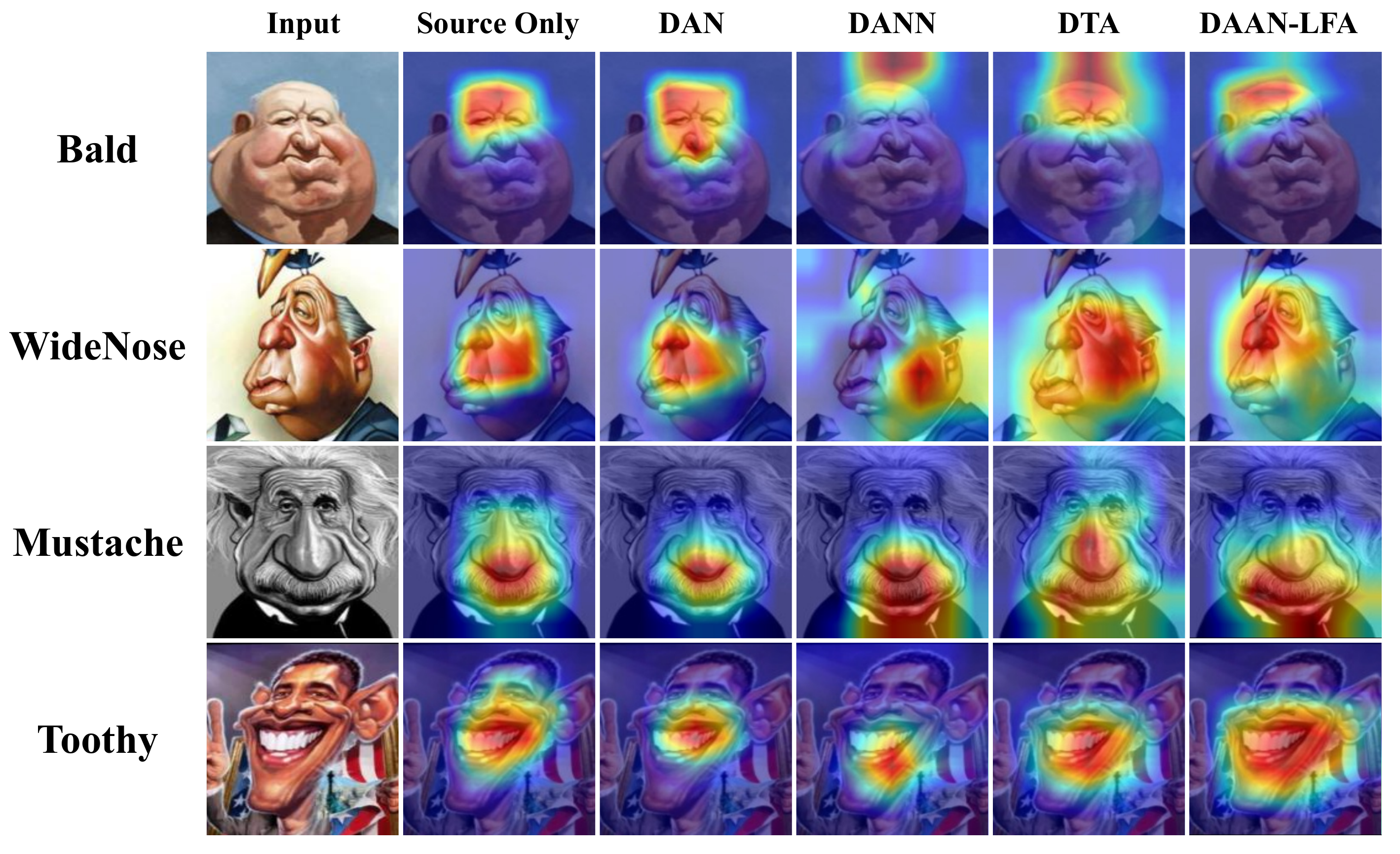}
    \caption{\label{fig:vis} The visualization of CAM for the proposed compared with the state-of-the-art methods on typical cases. Obviously, the proposed DAAN-LFA consistently generates high quality CAMs of accurate positions and active region sizes.}
\end{figure}

The CAM reveals the attention of the network when it makes decisions on a certain class. We visualize the CAMs for the proposed method compared with the state-of-the-art methods on the test set in Fig. \ref{fig:vis}. As the face attributes are often spatial-related, more precise attention regions for the specific attribute indicates higher predict performance. It is suggested by the figure that DAAN-LFA generates consistently best region activation through the cases. (More visualization results for DAAN is shown in \emph{Supplementary Material}.)

\section{Conclusions}

In this paper, we extend the 'WebCaricature' with fifty intrinsic face attributes to prepare the WebCariA dataset, for boosting the research in caricature attribute recognition. Then, to solve the problem of lacking annotated caricature attributes, we propose an unsupervised domain adaptation framework, \ie, DAAN, to estimate the caricature attributes by training the network with only given the face attribute labels. DAAN integrated with both inter- and intra-domain consistency learning paradigms. Specifically, in DAAN, an image-to-image translator and a label-consistency learning constrain are contained to fill the gap between photos and caricatures. To align the domain between the transferred images and original images in one certain domain, we propose the attribute-aware attention-consistency learning constrain, which is more efficient than the feature consistency constrain. The experiments show our framework can outperform the state-of-the-art methods by a reasonable margin.

\section*{Acknowledgement}

This work is supported in part by National Science Foundation of China under Grant No. 61806092, and in part by Jiangsu Natural Science Foundation under Grant No. BK20180326.

\clearpage
\bibliographystyle{splncs04}
\bibliography{0429}

\begin{thebibliography}{10}
\providecommand{\url}[1]{\texttt{#1}}
\providecommand{\urlprefix}{URL }
\providecommand{\doi}[1]{https://doi.org/#1}

\bibitem{abaci2015matching}
Abaci, B., Akgul, T.: Matching caricatures to photographs. Signal, Image and
  Video Processing  \textbf{9}(1),  295--303 (2015)

\bibitem{abdulnabi2015multi}
Abdulnabi, A.H., Wang, G., Lu, J., Jia, K.: Multi-task cnn model for attribute
  prediction. IEEE Transactions on Multimedia  \textbf{17}(11),  1949--1959
  (2015)

\bibitem{brennan2007caricature}
Brennan, S.E.: Caricature generator: The dynamic exaggeration of faces by
  computer. Leonardo  \textbf{40}(4),  392--400 (2007)

\bibitem{cao2018carigans}
Cao, K., Liao, J., Yuan, L.: Carigans: Unpaired photo-to-caricature
  translation. ACM Transactions on Graphics  \textbf{37}(6), ~244 (2018)

\bibitem{chen2017deeplab}
Chen, L.C., Papandreou, G., Kokkinos, I., Murphy, K., Yuille, A.L.: Deeplab:
  Semantic image segmentation with deep convolutional nets, atrous convolution,
  and fully connected crfs. IEEE transactions on pattern analysis and machine
  intelligence  \textbf{40}(4),  834--848 (2017)

\bibitem{csurka2017domain}
Csurka, G.: Domain adaptation for visual applications: A comprehensive survey.
  arXiv preprint arXiv:1702.05374  (2017)

\bibitem{ding2018deep}
Ding, H., Zhou, H., Zhou, S.K., Chellappa, R.: A deep cascade network for
  unaligned face attribute classification. In: Thirty-Second AAAI Conference on
  Artificial Intelligence (2018)

\bibitem{ehrlich2016facial}
Ehrlich, M., Shields, T.J., Almaev, T., Amer, M.R.: Facial attributes
  classification using multi-task representation learning. In: Proceedings of
  the IEEE Conference on Computer Vision and Pattern Recognition Workshops. pp.
  47--55 (2016)

\bibitem{dann}
Ganin, Y., Lempitsky, V.: Unsupervised domain adaptation by backpropagation.
  In: International conference on machine learning. pp. 1180--1189 (2015)

\bibitem{geng2013facial}
Geng, X., Yin, C., Zhou, Z.H.: Facial age estimation by learning from label
  distributions. IEEE transactions on pattern analysis and machine intelligence
   \textbf{35}(10),  2401--2412 (2013)

\bibitem{he2016deep}
He, K., Zhang, X., Ren, S., Sun, J.: Deep residual learning for image
  recognition. In: Proceedings of the IEEE conference on computer vision and
  pattern recognition. pp. 770--778 (2016)

\bibitem{he2017adaptively}
He, K., Wang, Z., Fu, Y., Feng, R., Jiang, Y.G., Xue, X.: Adaptively weighted
  multi-task deep network for person attribute classification. In: Proceedings
  of the 25th ACM international conference on Multimedia. pp. 1636--1644. ACM
  (2017)

\bibitem{hoffman2018cycada}
Hoffman, J., Tzeng, E., Park, T., Zhu, J.Y., Isola, P., Saenko, K., Efros, A.,
  Darrell, T.: Cycada: Cycle-consistent adversarial domain adaptation. In:
  Proceedings of the 35th International Conference on Machine Learning (2018)

\bibitem{huo2017webcaricature}
Huo, J., Li, W., Shi, Y., Gao, Y., Yin, H.: Webcaricature: a benchmark for
  caricature recognition. arXiv preprint arXiv:1703.03230  (2017)

\bibitem{isola2017image}
Isola, P., Zhu, J.Y., Zhou, T., Efros, A.A.: Image-to-image translation with
  conditional adversarial networks. In: Proceedings of the IEEE conference on
  computer vision and pattern recognition. pp. 1125--1134 (2017)

\bibitem{NIPS20083499}
Jacob, L., philippe Vert, J., Bach, F.R.: Clustered multi-task learning: A
  convex formulation. In: Koller, D., Schuurmans, D., Bengio, Y., Bottou, L.
  (eds.) Advances in Neural Information Processing Systems 21, pp. 745--752.
  Curran Associates, Inc. (2009),
  \url{http://papers.nips.cc/paper/3499-clustered-multi-task-learning-a-convex-formulation.pdf}

\bibitem{kim2019u}
Kim, J., Kim, M., Kang, H., Lee, K.H.: U-gat-it: Unsupervised generative
  attentional networks with adaptive layer-instance normalization for
  image-to-image translation. In: International Conference on Learning
  Representations (2019)

\bibitem{klare2012towards}
Klare, B.F., Bucak, S.S., Jain, A.K., Akgul, T.: Towards automated caricature
  recognition. In: 2012 5th IAPR International Conference on Biometrics (ICB).
  pp. 139--146. IEEE (2012)

\bibitem{kumar2012learning}
Kumar, A., Daume~III, H.: Learning task grouping and overlap in multi-task
  learning. ICML  (2012)

\bibitem{dta}
Lee, S., Kim, D., Kim, N., Jeong, S.G.: Drop to adapt: Learning discriminative
  features for unsupervised domain adaptation. In: Proceedings of the IEEE
  International Conference on Computer Vision. pp. 91--100 (2019)

\bibitem{liu2015deep}
Liu, Z., Luo, P., Wang, X., Tang, X.: Deep learning face attributes in the
  wild. In: Proceedings of the IEEE international conference on computer
  vision. pp. 3730--3738 (2015)

\bibitem{dan}
Long, M., Cao, Y., Wang, J., Jordan, M.: Learning transferable features with
  deep adaptation networks. In: International conference on machine learning.
  pp. 97--105 (2015)

\bibitem{lu2017fully}
Lu, Y., Kumar, A., Zhai, S., Cheng, Y., Javidi, T., Feris, R.: Fully-adaptive
  feature sharing in multi-task networks with applications in person attribute
  classification. In: Proceedings of the IEEE Conference on Computer Vision and
  Pattern Recognition. pp. 5334--5343 (2017)

\bibitem{luo2012hierarchical}
Luo, P., Wang, X., Tang, X.: Hierarchical face parsing via deep learning. In:
  2012 IEEE Conference on Computer Vision and Pattern Recognition. pp.
  2480--2487. IEEE (2012)

\bibitem{luo2013deep}
Luo, P., Wang, X., Tang, X.: A deep sum-product architecture for robust facial
  attributes analysis. In: Proceedings of the IEEE International Conference on
  Computer Vision. pp. 2864--2871 (2013)

\bibitem{mauro1992caricature}
Mauro, R., Kubovy, M.: Caricature and face recognition. Memory \& Cognition
  \textbf{20}(4),  433--440 (1992)

\bibitem{perkins1975definition}
Perkins, D.: A definition of caricature and caricature and recognition. Studies
  in Visual Communication  \textbf{2}(1),  1--24 (1975)

\bibitem{rudd2016moon}
Rudd, E.M., G{\"u}nther, M., Boult, T.E.: Moon: A mixed objective optimization
  network for the recognition of facial attributes. In: European Conference on
  Computer Vision. pp. 19--35. Springer (2016)

\bibitem{Russo_2018_CVPR}
Russo, P., Carlucci, F.M., Tommasi, T., Caputo, B.: From source to target and
  back: Symmetric bi-directional adaptive gan. In: Proceedings of the IEEE
  Conference on Computer Vision and Pattern Recognition (CVPR) (June 2018)

\bibitem{mcd}
Saito, K., Watanabe, K., Ushiku, Y., Harada, T.: Maximum classifier discrepancy
  for unsupervised domain adaptation. In: Proceedings of the IEEE Conference on
  Computer Vision and Pattern Recognition. pp. 3723--3732 (2018)

\bibitem{smith2017federated}
Smith, V., Chiang, C.K., Sanjabi, M., Talwalkar, A.S.: Federated multi-task
  learning. In: Advances in Neural Information Processing Systems. pp.
  4424--4434 (2017)

\bibitem{tsai2018learning}
Tsai, Y.H., Hung, W.C., Schulter, S., Sohn, K., Yang, M.H., Chandraker, M.:
  Learning to adapt structured output space for semantic segmentation. In:
  Proceedings of the IEEE Conference on Computer Vision and Pattern
  Recognition. pp. 7472--7481 (2018)

\bibitem{valentine2016face}
Valentine, T., Lewis, M.B., Hills, P.J.: Face-space: A unifying concept in face
  recognition research. The Quarterly Journal of Experimental Psychology
  \textbf{69}(10),  1996--2019 (2016)

\bibitem{vaswani2017attention}
Vaswani, A., Shazeer, N., Parmar, N., Uszkoreit, J., Jones, L., Gomez, A.N.,
  Kaiser, {\L}., Polosukhin, I.: Attention is all you need. In: Advances in
  neural information processing systems. pp. 5998--6008 (2017)

\bibitem{vazquez2012unsupervised}
V{\'a}zquez, D., L{\'o}pez, A.M., Ponsa, D.: Unsupervised domain adaptation of
  virtual and real worlds for pedestrian detection. In: Proceedings of the 21st
  International Conference on Pattern Recognition (ICPR2012). pp. 3492--3495.
  IEEE (2012)

\bibitem{wang2015deeply}
Wang, X., Guo, R., Kambhamettu, C.: Deeply-learned feature for age estimation.
  In: 2015 IEEE Winter Conference on Applications of Computer Vision. pp.
  534--541. IEEE (2015)

\bibitem{wang2017multi}
Wang, Z., He, K., Fu, Y., Feng, R., Jiang, Y.G., Xue, X.: Multi-task deep
  neural network for joint face recognition and facial attribute prediction.
  In: Proceedings of the 2017 ACM on International Conference on Multimedia
  Retrieval. pp. 365--374. ACM (2017)

\bibitem{zhangposition}
Zhang, Y., Shen, W., Sun, L., Li, Q.: Position-squeeze and excitation module
  for facial attribute analysis. BMVC  (2018)

\bibitem{zhou2016learning}
Zhou, B., Khosla, A., Lapedriza, A., Oliva, A., Torralba, A.: Learning deep
  features for discriminative localization. In: Proceedings of the IEEE
  conference on computer vision and pattern recognition. pp. 2921--2929 (2016)

\bibitem{zhu2017unpaired}
Zhu, J.Y., Park, T., Isola, P., Efros, A.A.: Unpaired image-to-image
  translation using cycle-consistent adversarial networks. In: Proceedings of
  the IEEE international conference on computer vision. pp. 2223--2232 (2017)

\bibitem{zhu2014multi}
Zhu, Z., Luo, P., Wang, X., Tang, X.: Multi-view perceptron: a deep model for
  learning face identity and view representations. In: Advances in Neural
  Information Processing Systems. pp. 217--225 (2014)

\end{thebibliography}
\end{document}